\documentclass[10pt,twocolumn,letterpaper]{article}

\usepackage{cvpr}
\usepackage{times}
\usepackage{epsfig}
\usepackage{graphicx}
\usepackage{amsmath}
\usepackage{amssymb}

\usepackage[breaklinks=true,bookmarks=false]{hyperref}

\cvprfinalcopy 

\def\cvprPaperID{****} 

\begin{document}

\title{PNUNet: Anomaly Detection using Positive-and-Negative Noise based on \\ Self-Training Procedure}

\author{Masanari Kimura\\
Ridge-i inc.\\
{\tt\small mkimura@ridge-i.com}
}

\maketitle

\maketitle

\section*{Abstract}
We propose the novel framework for anomaly detection in images. Our new framework, PNUNet, is based on many normal data and few anomalous data. We assume that some noises are added to the input images and learn to remove the noise. In addition, the proposed method achieves significant performance improvement by updating the noise assumed in the inputs using a self-training framework. The experimental results for the benchmark datasets show the usefulness of our new anomaly detection framework.

\section{Introduction}
In recent years, many methods based on Deep Neural Networks (DNNs) for detecting anomalous areas in images have been studied \cite{schlegl2017unsupervised, baur2018deep, kimura2018semi, tian2019learning}. These studies are beginning to draw much attention as the industry develops.

There are the following requirements for anomaly detection methods in images that are expected to be used in factories: (1) High generalization performance that can handle unknown data, (2) Runtime speed that enables anomaly detection in real time, (3) Ease of learning network. Anomaly detection methods in images are widely used based on Auto Encoders and those based on (Generative Adversarial Networks) GANs. However, in most cases, the methods based on Auto Encoder does not satisfy requirement (1), and the method based on GANs does not satisfy (2) and (3). Auto-encoder based methods cannot detect anomalies clearly. On the other hand, GANs-based methods have to solve another optimization problem at the time of inference, so the inference speed is slow and learning of GANs is difficult.

\begin{figure}[tbp]
    \centering
    \includegraphics[scale=0.34]{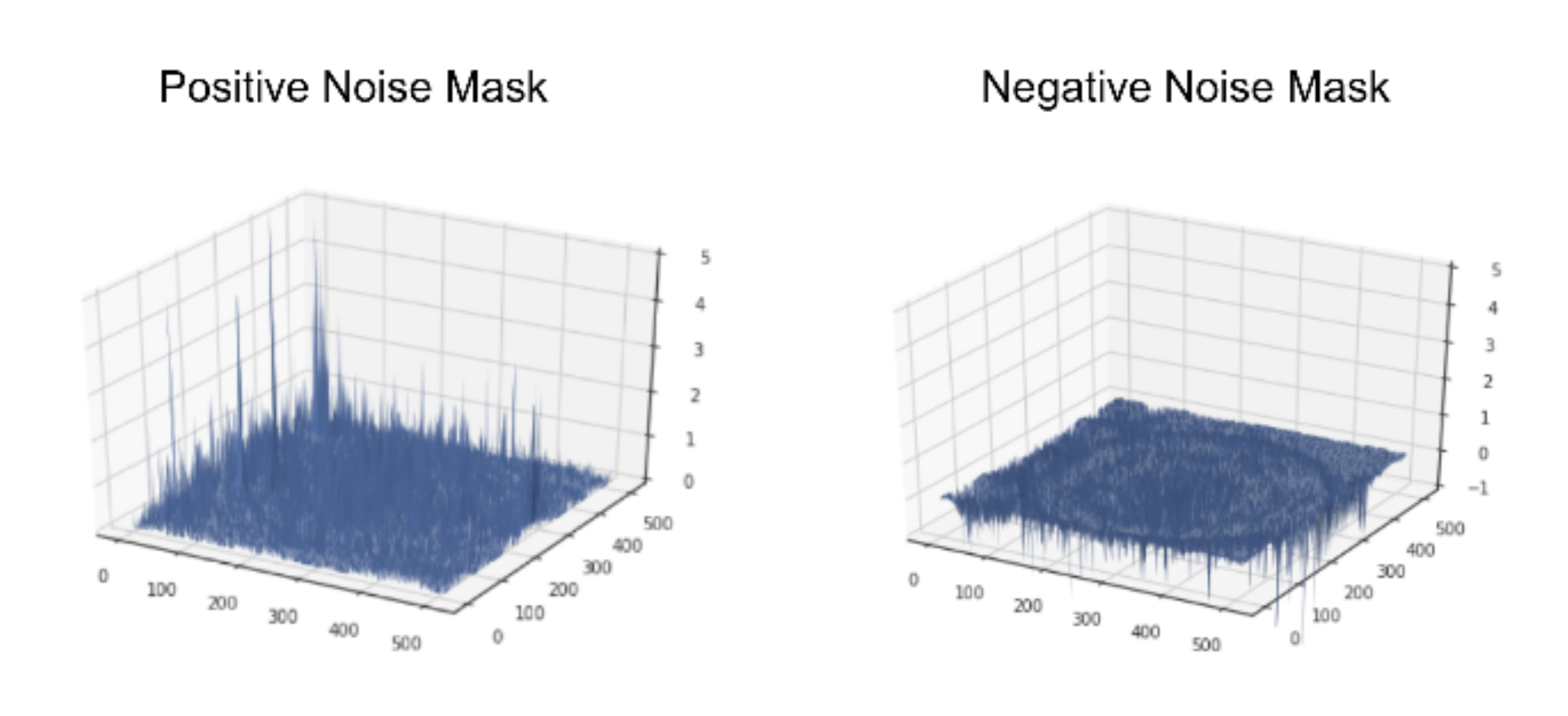}
    \caption{Positive-and-Negative Noise Mask. Positive Noise Mask generates noise that makes the abnormal area stand out. On the other hand, Negative Noise Mask suppresses noise that is always included in normal images.     \label{fig:pnmask}}
\end{figure}

To solve these problems, we propose a novel anomaly detection framework that meets all three requirements. Our proposed method is based on U-Net\cite{ronneberger2015u}, which has achieved great success in semantic segmentation task. Our main idea is to assume that noise is added to the input image, generate an image from which the noise is removed, and take the difference. Intuitively, it is desirable that the noise added to the image be close to the actual abnormal area. Therefore, we propose a method to approximate the distribution of noise used for learning to the actual abnormal noise by the self-training procedure. We call this pseudo-anomaly Positive-and-Negative Noise (see Figure \ref{fig:pnmask}).

In summary, our contributions are as follows.
\begin{itemize}
    \item{We propose a novel method to detect anomalous area by comparing the input image with the generated image from which the anomalous part has been removed.}
    \item{We introduce a new framework that updates the noise used for learning as a pseudo-anomaly using self-training.}
    \item{The experimental results for the benchmark datasets show the usefulness of our new anomaly detection framework.}
\end{itemize}

\begin{figure*}[t]
    \centering
    \includegraphics[scale=0.45]{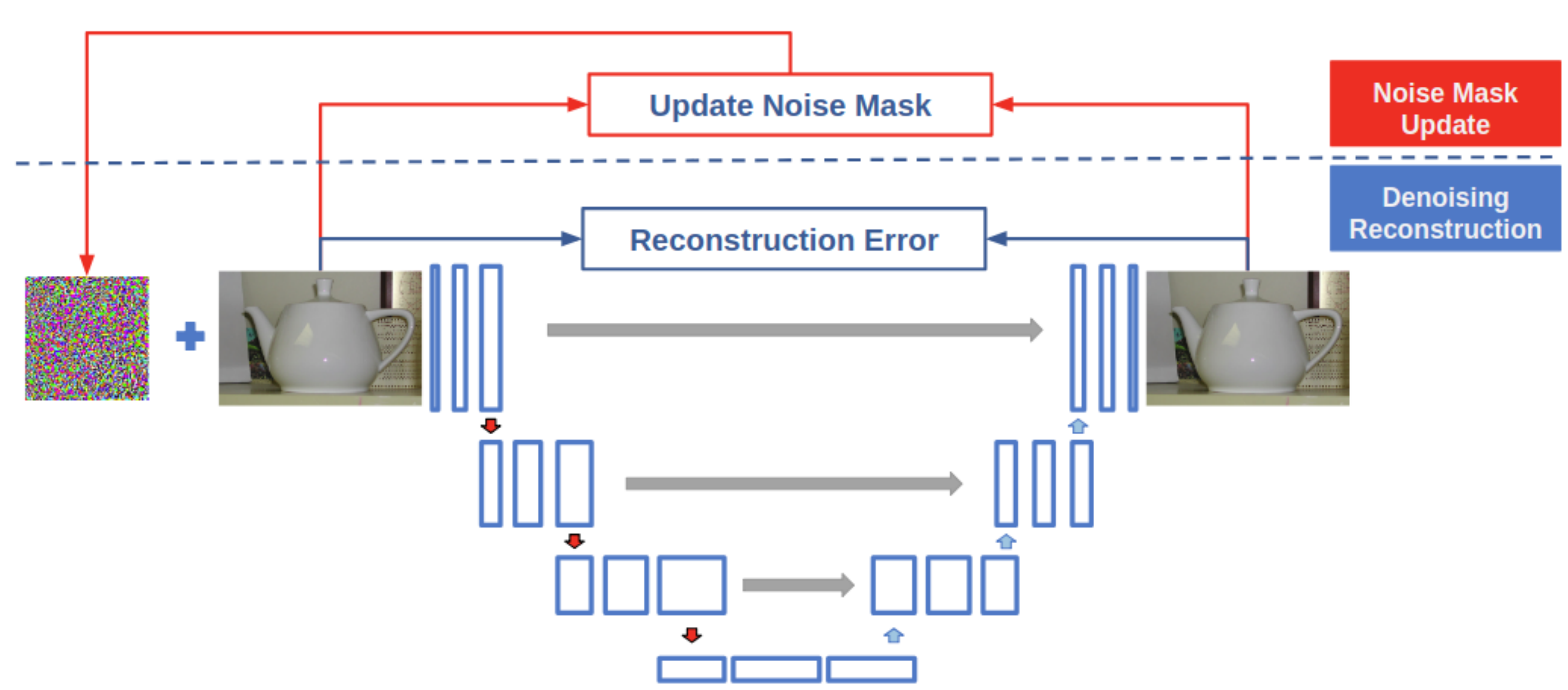}
    \caption{Overview of proposed method. he proposed method is divided into two components: Noise removal component and Noise mask update component.     \label{fig:network_overview}}
\end{figure*}

\begin{figure*}[t]
    \centering
    \includegraphics[scale=0.45]{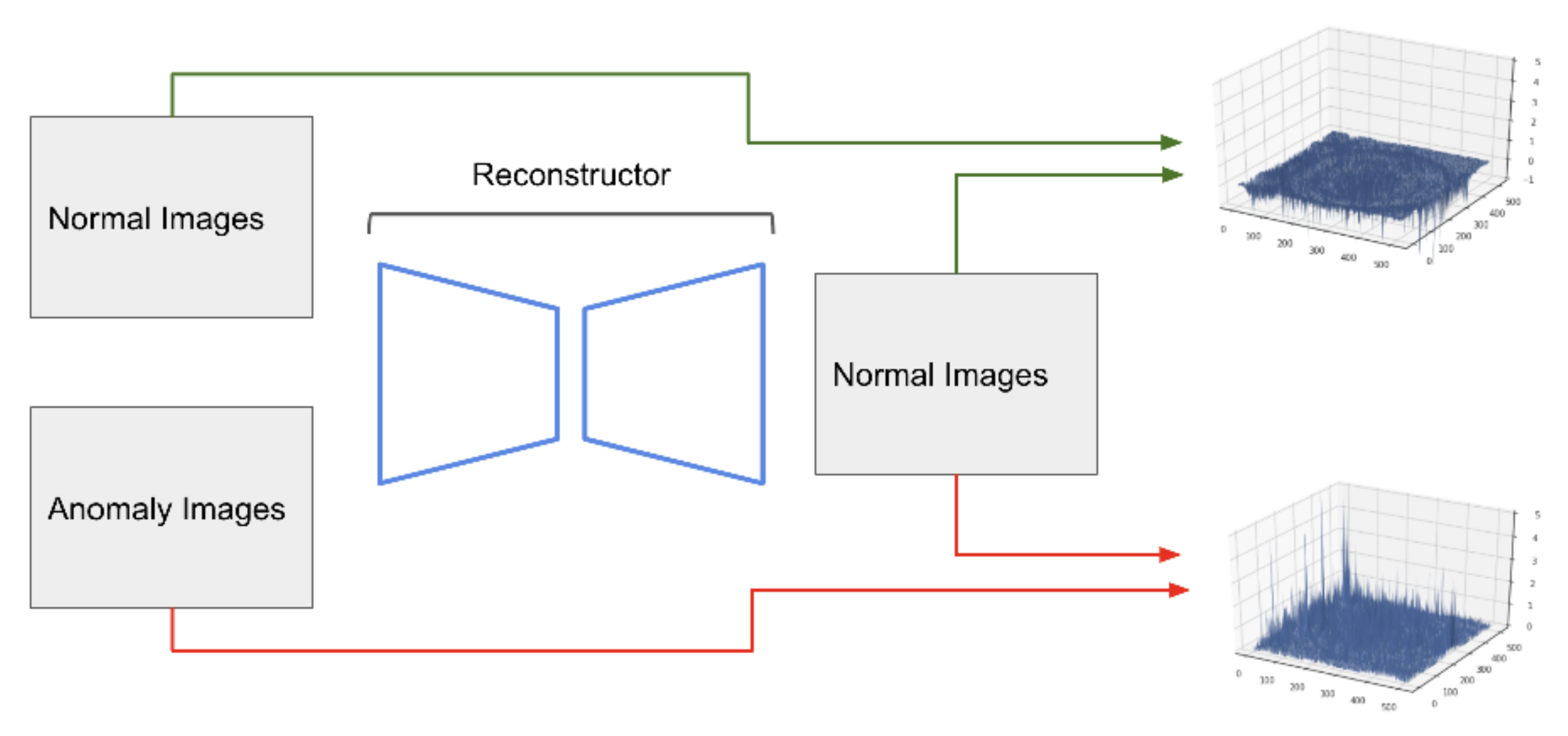}
    \caption{Noise mask update component. Positive Noise Mask is generated from an image containing an anomaly, and Negative Noise Mask is generated from a normal image.     \label{fig:mask_update}}
\end{figure*}

\section{Related Works}
The detection of anomaly areas in images is widely studied \cite{an2015variational, schlegl2017unsupervised, baur2018deep, kimura2018semi, tian2019learning}. The VAEs-based method finds anomaly areas by taking the difference between the input image and the reconstructed image. AnoGAN \cite{schlegl2017unsupervised} brought great progress to the field with a simple algorithm that only normal images are used for training a generator to model the distribution of normal images. With the trained generator $G$, if a given new query image $x$ is from the normal data distribution, noise $\hat{z}$ must exist in the latent space where $G(\hat{z})$ becomes identical to $x$. Furthermore, a method that improves AnoGAN has also been proposed \cite{kimura2018semi, zenati2018efficient}.

Although these GANs-based methods are powerful, they have the problem that inference takes a very long time. To address this, we propose a framework, called PNUNet, that enables fast inference while maintaining GANs-based detection accuracy.

\section{Proposed Method}
Figure \ref{fig:network_overview} shows an overview of the proposed method. The proposed method is roughly divided into two components.
\begin{itemize}
    \item{Noise removal component}
    \item{Noise mask update component}
\end{itemize}
We explain the details of each component below.

\begin{figure*}[t]
    \centering
    \includegraphics[scale=0.45]{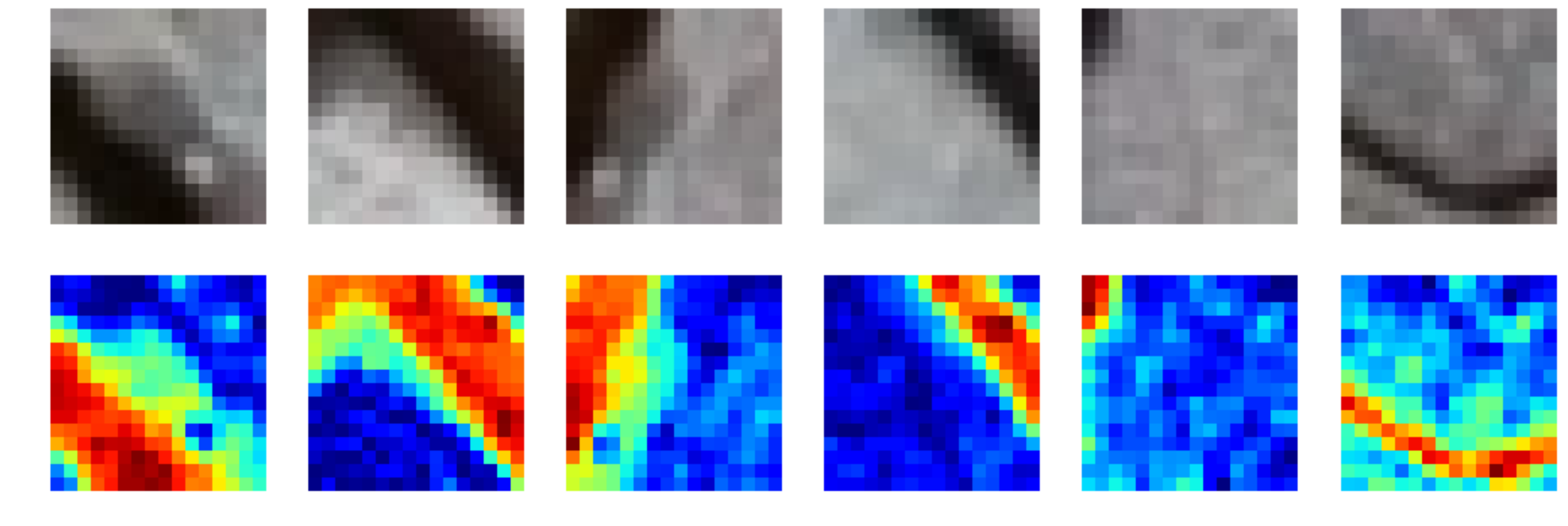}
    \caption{Experimental results for Bridge Crack Image Data. The first row is the inputs images and the second row is the differences between input images and generated images. \label{fig:bridge_crack}}
\end{figure*}

\begin{figure}[t]
    \centering
    \includegraphics[scale=0.33]{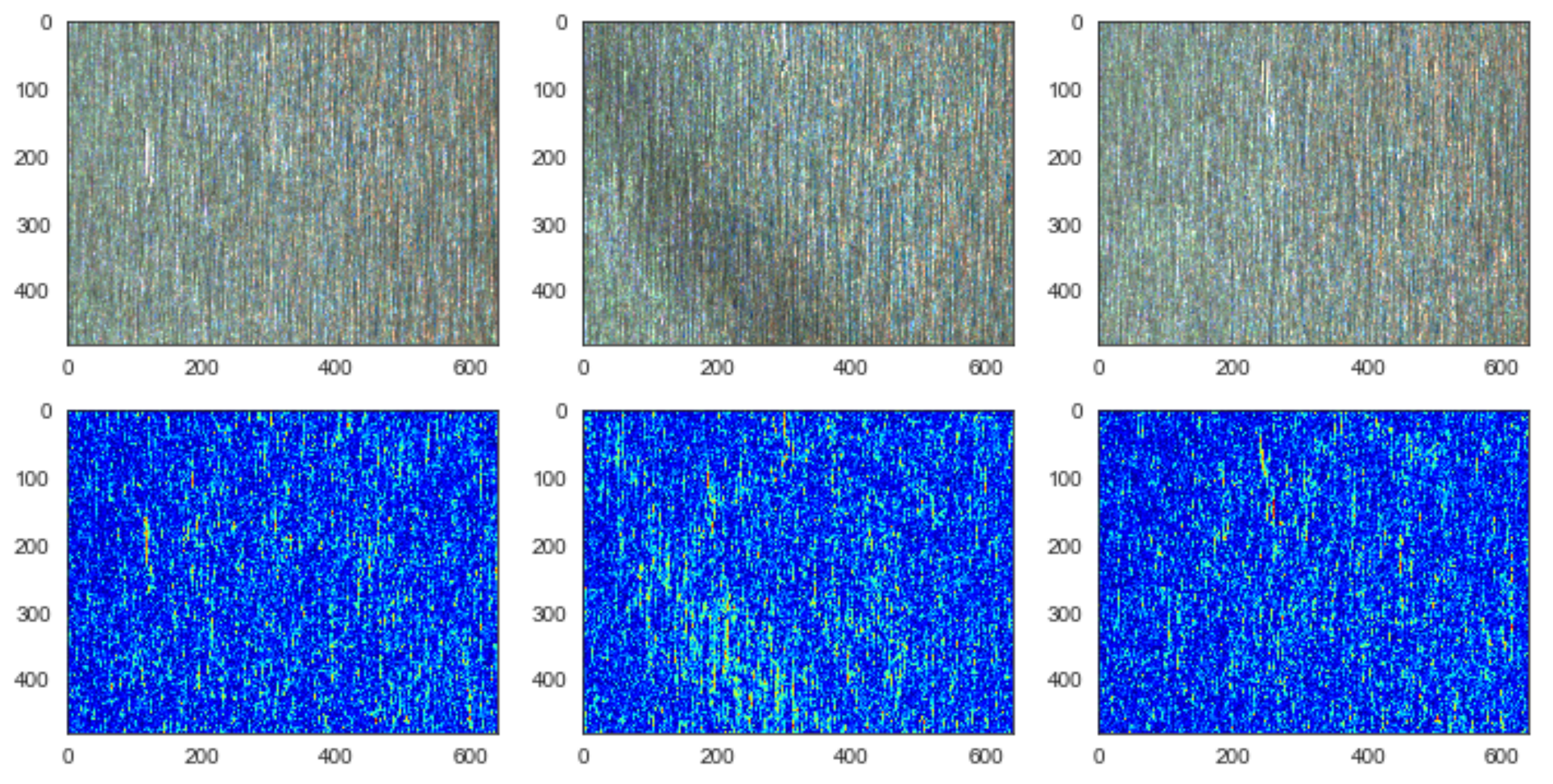}
    \caption{Experimental results for Micro Surface Defect Detection. The first row is the inputs images and the second row is the differences between input images and generated images. \label{fig:sdd}}
\end{figure}

\subsection{Anomaly Detection using Noise Removal Network}
We assume the noise $z$ obtained from the distribution $P(z)$ to the input image. The Noise removal component $f_r(\cdot)$ removes this noise from the image. The loss function is as follows:
\begin{equation}
    L = L_r(x, f_r(x + z)), \ \ z\sim{\mathbf{P}(z)},
\end{equation}
Here, $x$ is an input image, $z$ is a noise, and $f_r$ is a reconstructor. We use Structural SIMilarity (SSIM) as a reconstruction error $L_r$.

There is a problem that the detection accuracy of the anomaly area becomes low when the uniform noise distribution is used simply for the learning of the noise removal component $f_r(\cdot)$. It is assumed that this problem is due to the following two causes.
\begin{itemize}
    \item{It cannot be learned to remove noise similar to real anomaly areas (such as scratches and distortion).}
    \item{It is learned to remove noise (such as brightness) that is always included in normal images.}
\end{itemize}
To tackle this, we propose a new framework to learn noise masks simultaneously.

\subsection{Self-Training for Noise Updating}
Figure \ref{fig:mask_update} shows noise mask update component.
Positive Noise Mask is generated from an image containing an anomaly $x_p$, and Negative Noise Mask is generated from a normal image $x_n$.

The positive noise $z_p$ and the negative noise $z_n$ are as follows:
\begin{eqnarray}
    z_p &= z\cdot{|x_p - f_r(x_p)|}, \\
    z_n &= 1 - z\cdot{|x_n - f_r(x_n)|},
\end{eqnarray}
The positive noise $z_p$ emphasizes the area that is likely to be anomaly, and the negative noise $z_n$ suppresses the area that is likely to be normal.

In our framework, masks are generated at regular intervals during the learning of the reconstructor. In other words, in our proposed method, the reconstructor and the noise mask grow simultaneously.

\section{Experimental Results}
In the experiments, the noise distribution of the proposed method is a uniform distribution, and the noise masks are updated every $1,000$ iteration.

Figure \ref{fig:bridge_crack} shows the experimental results for Bridge Crack Image Data \footnote{$https://github.com/maweifei/Bridge_Crack_Image_Data$}, which contains 50,000 train images and 5,000 validation images. The proposed method finds out the anomaly area by taking the difference between the input image and the result of reconstructing the input image with the reconstructor.

Figure \ref{fig:sdd} shows the experimental results for Micro Surface Defect Detection (MSDD) \cite{song2013micro}. The defect image contains the micro defect object and the clutter background. The proposed method can also visualize very small anomaly areas in the $480\times{640}$ high-resolution image. In addition, Table \ref{table:runtime} shows the comparison of inference speed against the MSDD dataset. The proposed method is much faster than the GANs-based comparison method.

\section{Conclusion and Discussion}
We proposed a novel framework to detect abnormal areas in the image.
Our proposed method generates an image in which an anomalous area is removed from an input image by a denoising network, and achieves anomaly detection by taking the difference between the generated image and the input image. Our proposed method solves the problems with the conventional method. The experimental results show the usefulness of our framework in the visual inspection task.

\begin{table}[t]
\centering
\caption{Comparison of AnoGAN and PNUNet inference speed for MSDD dataset. \label{table:runtime}}
\begin{tabular}{l|c}
\hline
Method & Inference time per image {[}sec{]} \\ \hline
AnoGAN & 47.082                             \\
PNUNet & 0.035                          \\ \hline
\end{tabular}
\end{table}

\bibliographystyle{ieee_fullname}

\begin{thebibliography}{1}\itemsep=-1pt

\bibitem{an2015variational}
Jinwon An and Sungzoon Cho.
\newblock Variational autoencoder based anomaly detection using reconstruction
  probability.
\newblock {\em Special Lecture on IE}, 2:1--18, 2015.

\bibitem{baur2018deep}
Christoph Baur, Benedikt Wiestler, Shadi Albarqouni, and Nassir Navab.
\newblock Deep autoencoding models for unsupervised anomaly segmentation in
  brain mr images.
\newblock In {\em International MICCAI Brainlesion Workshop}, pages 161--169.
  Springer, 2018.

\bibitem{kimura2018semi}
Masanari Kimura and Takashi Yanagihara.
\newblock Semi-supervised anomaly detection using gans for visual inspection in
  noisy training data.
\newblock {\em arXiv preprint arXiv:1807.01136}, 2018.

\bibitem{ronneberger2015u}
Olaf Ronneberger, Philipp Fischer, and Thomas Brox.
\newblock U-net: Convolutional networks for biomedical image segmentation.
\newblock In {\em International Conference on Medical image computing and
  computer-assisted intervention}, pages 234--241. Springer, 2015.

\bibitem{schlegl2017unsupervised}
Thomas Schlegl, Philipp Seeb{\"o}ck, Sebastian~M Waldstein, Ursula
  Schmidt-Erfurth, and Georg Langs.
\newblock Unsupervised anomaly detection with generative adversarial networks
  to guide marker discovery.
\newblock In {\em International Conference on Information Processing in Medical
  Imaging}, pages 146--157. Springer, 2017.

\bibitem{song2013micro}
Kechen Song and Yunhui Yan.
\newblock Micro surface defect detection method for silicon steel strip based
  on saliency convex active contour model.
\newblock {\em Mathematical Problems in Engineering}, 2013, 2013.

\bibitem{tian2019learning}
Kai Tian, Shuigeng Zhou, Jianping Fan, and Jihong Guan.
\newblock Learning competitive and discriminative reconstructions for anomaly
  detection.
\newblock {\em arXiv preprint arXiv:1903.07058}, 2019.

\bibitem{zenati2018efficient}
Houssam Zenati, Chuan~Sheng Foo, Bruno Lecouat, Gaurav Manek, and
  Vijay~Ramaseshan Chandrasekhar.
\newblock Efficient gan-based anomaly detection.
\newblock {\em arXiv preprint arXiv:1802.06222}, 2018.

\end{thebibliography}

\end{document}